\pdfoutput=1

\documentclass[11pt]{article}

\usepackage[]{acl}

\usepackage{times}
\usepackage{latexsym}

\usepackage[T1]{fontenc}

\usepackage[utf8]{inputenc}

\usepackage{microtype}

\usepackage{graphicx}
\usepackage{multicol}
\usepackage{multirow}
\usepackage{amsmath}

\usepackage{color}

\title{Leveraging Graph-based Cross-modal Information Fusion for Neural Sign Language Translation}

\author{Jiangbin Zheng\textsuperscript{1,2,3}, Siyuan Li\textsuperscript{1,2,3}, \textbf{Cheng Tan\textsuperscript{1,2,3}, Chong Wu\textsuperscript{4}, Yidong Chen\textsuperscript{5}, Stan Z. Li\textsuperscript{2,3}} \\
    \textsuperscript{1}Zhejiang University \\
    \textsuperscript{2}AI Lab, School of Engineering, Westlake University \\
    \textsuperscript{3}Institute of Advanced Technology, Westlake Institute for Advanced Study \\
    \textsuperscript{4}City University of Hong Kong, Kowloon, Hong Kong SAR \\
    \textsuperscript{5}Department of Artificial Intelligence, School of Informatics, Xiamen University \\
}

\begin{document}
\maketitle
\begin{abstract}
\textbf{Sign Language} (SL), as the mother tongue of the deaf community, is a special visual language that most hearing people cannot understand. In recent years, neural \textbf{Sign Language Translation} (SLT), as a possible way for bridging communication gap between the deaf and the hearing people, has attracted widespread academic attention. We found that the current mainstream end-to-end neural SLT models, which tries to learning language knowledge in a weakly supervised manner, could not mine enough semantic information under the condition of low data resources. Therefore, we propose to introduce additional word-level semantic knowledge of sign language linguistics to assist in improving current end-to-end neural SLT models. Concretely, we propose a novel neural SLT model with multi-modal feature fusion based on the dynamic graph, in which the cross-modal information, i.e. text and video, is first assembled as a dynamic graph according to their correlation, and then the graph is processed by a multi-modal graph encoder to generate the multi-modal embeddings for further usage in the subsequent neural translation models. To the best of our knowledge, we are the first to introduce graph neural networks, for fusing multi-modal information, into neural sign language translation models. Moreover, we conducted experiments on a publicly available popular SLT dataset \textit{RWTH-PHOENIX-Weather-2014T}. and the quantitative experiments show that our method can improve the model.
\end{abstract}

\section{Introduction}

\textbf{Sign Language} (SL), as the mother tongue of the deaf community, is a special visual language that most hearing people cannot understand. Thus, \textbf{Sign Language Translation} (SLT), as an application that may bridge the communication between the deaf and the hearing people, has recently attracted the attention of researchers \cite{camgoz2018neural,ko2019neural,camgoz2020sign,camgoz2020multi,zheng2020improved,zheng2021enhancing}. To be precise, the goal of SLT is to convert the input continuous sign language video into its semantically equivalent spoken language translation \cite{camgoz2018neural}.

Up to now, mainstream SLT work is based on deep neural network framework, while due to the lack of training datasets, the current neural SLT tasks are still looked on as weakly supervised tasks in low-resource scenarios \cite{koller2019weakly}. Actually, we could roughly divide the current models into the following three categories:

The first approach decomposes the SLT problem into two stages. For example, Camgoz et al \cite{camgoz2018neural} first used a Continuous Sign Language Recognition (CSLR) method \cite{koller2017re} to obtain the sign language gloss sequence and then used a \textbf{Neural Machine Translation} (NMT) model \cite{luong2015effective,vaswani2017attention} to translate the glosses into the spoken sentences \cite{camgoz2018neural}.

The second approach focuses on direct learning from sign language video representations to spoken translation without intermediate layer representations \cite{camgoz2018neural,ko2019neural}. Theoretically, with sufficient datasets and complex network architectures, these models can achieve excellent end-to-end neural SLT without the use of any manually annotated annotation information. However, due to the lack of a fully supervised approach to guide deeper understanding of sign language, this approach performs significantly lower than other approaches on the currently available low-resource datasets.

The third approache is the recently proposed Transformer-based \cite{vaswani2017attention} end-to-end approach for joint training. Camgoz et al \cite{camgoz2020multi} train the shared encoder with the aid of a CSLR model, introducing intermediate supervision at the sign language lexical level, which helps the network learn more meaningful feature representations in the temporal and spatial domains of sign language, but does not restrict the information transfer to the auto-regressive SLT Transformer decoder. Although the method makes new improvements in the NMT module, its shallow spatial feature module still follows the previous approach.

Under the condition of low data resources, it is difficult for the end-to-end SLT architectures to mine enough implicit semantic information only with the help of the deep network, so their performances are generally poor. Actually, we found that in earlier non-end-to-end multi-stage models, SLT models relied on recognition models to generate sign glosses as intermediate supervision to enhance the effect and achieved SOTA results \cite{camgoz2018neural,camgoz2020sign}. Furthermore, previous work \cite{koller2017re,koller2019weakly,cui2019deep} has shown that introducing additional information into the end-to-end model can improve the overall translation performance. Therefore, we argue that based on the end-to-end training mode, the text information, which was originally used in the middle layer, can be used as the input of the embedded layer as auxiliary semantic knowledge and help improving the translation performance. Moreover, considering that the current sign language translation models only use video features as the input of the embedding layer, introducing additional text information, i.e. glosses, as auxiliary features will lead to new challenges, such as 1) how to fuse features from different modalities, and 2) how to extract and use the correlation between cross-modal information.

Therefore, to break the translation bottleneck of sign language in low-resource conditions and to be able to apply glosses as external sign language expertise to a neural SLT model, we propose a novel neural SLT model with multi-modal feature fusion based on the dynamic graph, as shown in Fig. 1. In this model we propose to use the representation of graph network structure to express complex non-Euclidean relationships to solve the fusion problem of multi-modal features, which is the first work to introduce the concept of graph network into the neural SLT model. In the initialization stage of the model, we a pseudo-labeling form instead of the manual labeling form and implement the semantic segmentation of sign language with unsupervised or semi-supervised algorithms to calculate the clustering relationships of video frame sequences and the alignment relationships with text words. Based on the multi-modal mapping relationships, a graph-based sign language feature encoder can be further constructed. During the training process, we further optimize the alignment relations by a dynamic iterative strategy.

The contributions of this paper can be summarized as:

1. For the first time, additional information other than video information is introduced and represented in a multi-modal form in the encoding of a neural SLT model. For this purpose, we introduced the concept of graph neural SLT model for the first time. The multi-modal information of sign language was successfully fused using the construction of graph neural networks. There is no precedent of applying graph coding to a SLT model in previous work. 

2. Sign language segmentation was implemented using a semi-supervised clustering algorithm and applied for the first time to a SLT task. Sign language segmentation was mainly used to map the alignment relationships of multi-modal sequences, assist graph network construction and multi-modal feature fusion. In addition, an iterative alignment method for pseudo-label sequences is introduced in the training process to dynamically align visual and text sequences to realign the graph structure in an iterative update.


\begin{figure}
  \centering
  \includegraphics[width=0.9\linewidth]{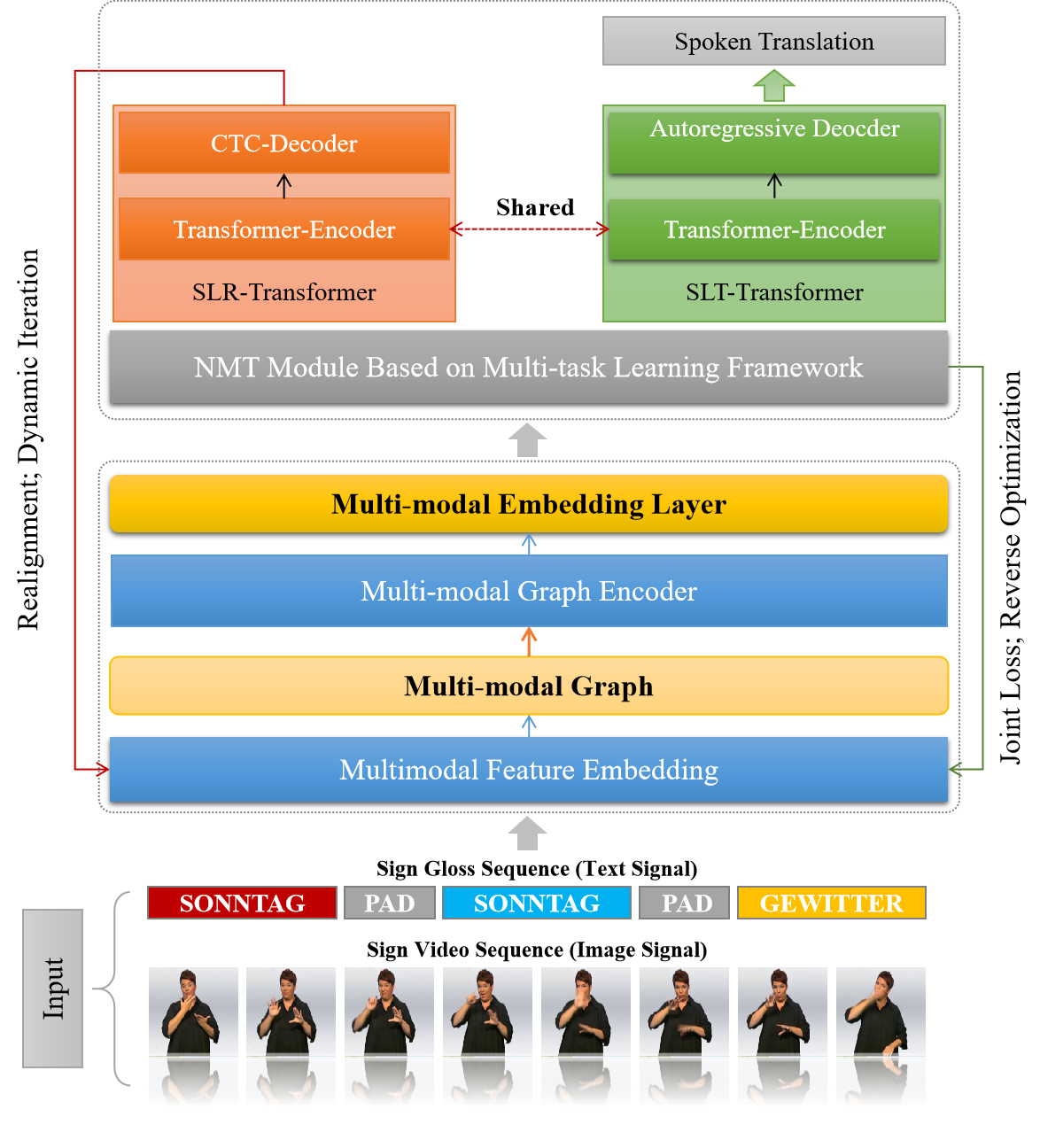}
  \caption{The overall construction diagram of our proposed graphical neural sign language translation model.}
\end{figure}

\section{Related Work}
The early sign language automation tasks were mainly for sign language recognition \cite{kamal2019technical}. According to the types of sign language recognition, sign language recognition systems can usually be classified into three categories, i.e., Finger-spelling Recognition, Isolated Word Recognition, and Continuous Sign Sentences Recognition. Initially, due to technical limitations, research on sign language recognition was focused on lexical-level Finger-spelling Recognition \cite{peng2009chinese,ji20163d,pan2018sign} and Isolated Word Recognition \cite{gao2004chinese,fang2006large,wang2015fast,wang2016isolated,zhang2016chinese,zhuang2017towards,huang2018novel,liang20183d}. For real-life communication between the hearing and the deaf people, the later emerging Continuous Sign Sentences Recognition \cite{cui2019deep,niu2020stochastic,cheng2020fully,zhou2020spatial,guo2019dense,pu2019iterative,de2019spatial} is more useful but also more difficult.

However, compared to sign language recognition, neural SLT is more practical, which further focuses on the deeper natural language properties of sign language such as word order and semantics. Therefore, it is currently a more advanced sign language task. Camgoz et al. (2018) \cite{camgoz2018neural} first implemented a neural SLT model that can generate spoken translations from sign language videos. They proposed for the first time to treat the SLT problem as an NMT problem. This work was pioneering in the field of neural SLT and laid an important foundation for the subsequent work. To evaluate the performance of the SLT, they also released the first publicly available SLT dataset, RWTH-PHOENIX-Weather 2014T.

Later, Ko et al. (2018) \cite{ko2019neural} proposed an SLT system based on human keypoint estimation. The obtained human keypoint vector was normalized by the mean and standard deviation of the keypoints and used as input to a Seq2Seq translation model. Orbay et al. (2020) \cite{orbay2020neural} proposed to explore semi-supervised tokenization using various methods such as adversarial, multi-task, and transfer learning without adding additional burden of annotation. Camgoz et al. (2020) \cite{camgoz2020multi} proposed a multichannel SLT architecture initially overcoming the reliance on sign gloss annotation by considering multiple asynchronous information channel expression mechanisms. Camgoz et al. (2020) \cite{camgoz2020sign} then applied Transformer for the first time to sign language recognition and translation tasks. This model is the current state-of-the-art baseline for jointly training CSLR and SLT modules in an end-to-end manner without any real temporal information while alleviating the problem of interdependent sequence learning.

Overall, compared to the much earlier emergence of sign language recognition tasks, research on neural SLT started late and is still at a rudimentary level. Meanwhile, the lack of dataset resources and the complexity of sign language itself make the SLT task a high challenge in sign language research.

\begin{figure}[h]
  \centering
  \includegraphics[width=\linewidth]{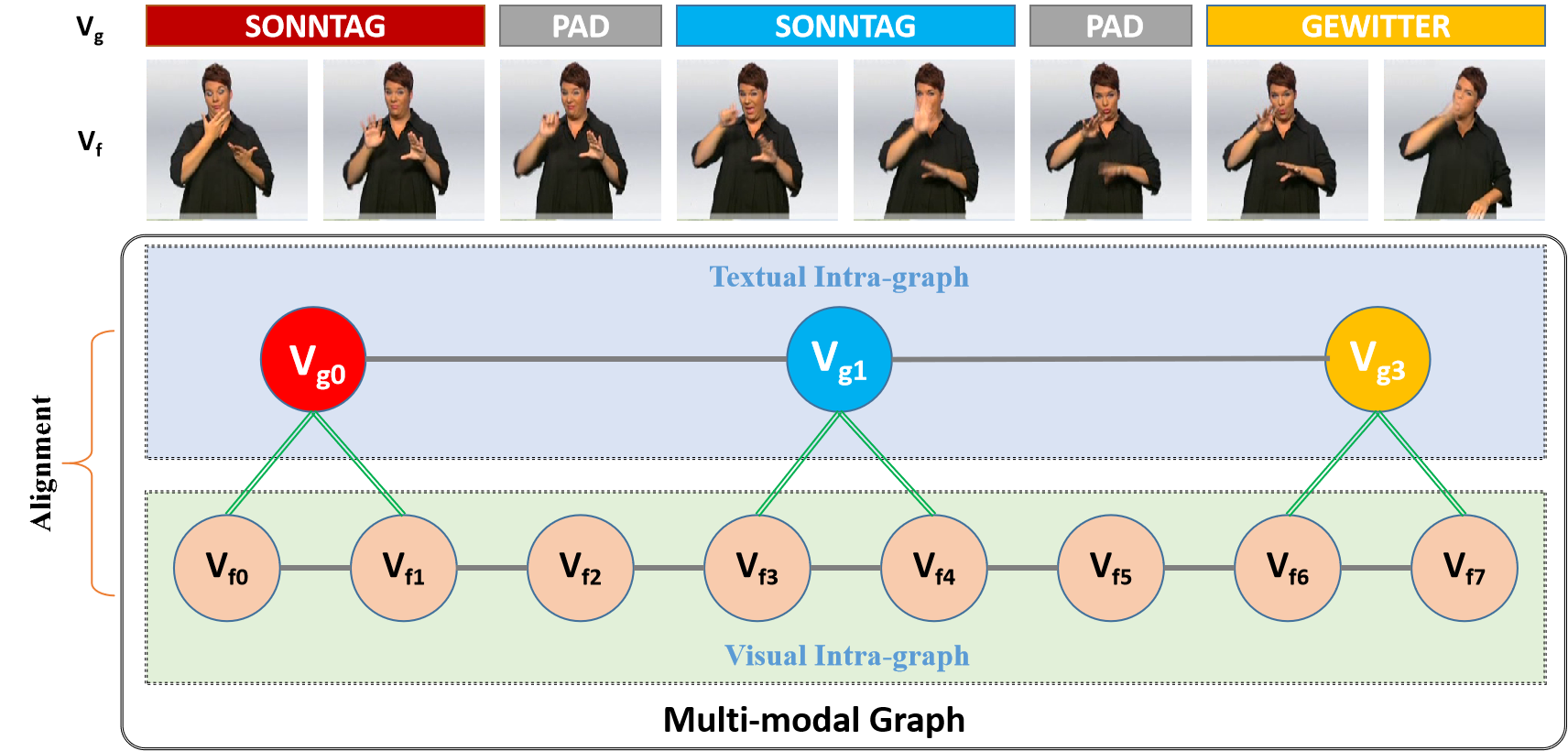}
  \caption{A sample of aligning textual and visual nodes in a multi-modal graph.}
\end{figure}

\section{Our Proposed Methods}
Fig. 1 shows the overall architecture diagram of our proposed model. As an improvement to the feature extraction module, we additionally introduce a Textual Embedding module based on the Spatial Embedding module to align sequences and fuse multi-modal information in the form of a graph network\cite{wu2022star,xia2022towards}. The input features are first fed to a \textbf{Multi-Modal Embedding} module and a \textbf{Multi-modal Graph Encoder} module sequentially to fuse the different modal features, and then fed to a \textbf{NMT} module based on the multitask learning framework where multiple Transformer modules are jointly trained in an end-to-end manner. Besides, we introduce a mode to dynamically update the construction of multi-modal graphs using pseudo-label iterative alignment during training for further optimization. The multi-modal graph will be reconstructed when the parameters of the CSLR module obtain better parameters. The reconstructing process is the same as the initialization process. In the next subsections, we will describe our modules in detail.

\subsection{Multi-modal Sequence Alignment}
In the initialization phase, we initially obtain the temporal relationships of the generated text sequences as well as the frame sequences based on the CSLR model. After obtaining the graph alignment relationships, we perform the initial fusion of graph features using a multi-modal graph encoder. In the subsequent steps, the parameters of the CSLR module are continuously updated after iterative training and back-propagation. If the word error rate further decreases and better CSLR performance is obtained, the graph network alignment relations are dynamically reconstructed. Next, we will introduce the process of graph construction in detail.

\subsubsection{Multi-modal Graph Definition}
The graph we define is an undirected graph, formalized as $G = (V, E)$, where $V$ is a \textbf{Node Set} and $E$ is an \textbf{Edge Set}.

In \textbf{Node Set} $V$, (1) we treat all the words of the sign gloss sequence as independent textual nodes $V_{fi}$. For example, in Fig. 2 the multi-modal graph contains a total of 3 textual nodes, each corresponding to a word in the input sentence; (2) we treat all video frames as independent visual nodes $V_{gj}$. For example, in Fig. 2 the multi-modal graph contains a total of 8 visual nodes, each corresponding to a frame in the input video.

In \textbf{Edge Set} $E$, (1) any two nodes in the same modality are connected by an \textbf{Intra-Modal Edge}. For example, the connection between one visual (or textual) node and another visual (or textual) node; (2) any textual node representing sign gloss and the corresponding visual node are connected by \textbf{Inter-Modal Edges}.As shown in Fig. 2, only few of the nodes such as $V_{f1}$ and $V_{g0}$ are interconnected by \textbf{Inter-Modal Edges}.

\subsubsection{Pseudo-Label Sequence Acquisition}
Next, the alignment relationship between $V_{fi}$ and $V_{gj}$ has to been calculated. 

CSLR models usually consist of visual feature modules and CTC modules sequentially combined. A pre-trained CSLR can output sign gloss sequences with low word error rate (which we call \textit{Pseudo-Sign Gloss Sequences}) where the length of the sign gloss sequence is usually much smaller than the length of the input video frames, due to the CTC layer optimizing the candidate gloss sequences (removing spaces, repetitive words, etc.) . We can obtain the alignment between the original sequence of the sign glosses (which we call \textit{Pseudo-Primitive Sign Gloss Sequence}) and the video frames before processing by the CTC layer, where the length of the original sequence is equal to the length of the input video frames.

Suppose that the input receives a video frame sequence with N frames, and its corresponding visual node sequence is denoted as $V_{f}=\{V_{f0},V_{f1},... ,V_{fi},... ,V_{fN}\}$. By loading the complete pre-trained CSLR model, we can obtain the alignment sequence of \textit{Pseudo-Sign Glosses} for $M$ words as $Vg_{ctc}=\{Vg_{ctc0},...,Vg_{ctci},...,Vg_{ctcM}\}$. And the \textit{Pseudo-Primitive Sign Gloss Sequence} features obtained without CTC layer processing are denoted as $P=\{P_{0},P_{1},... ,P_{i},... ,P_{N}\}$ ($Pi \in \{GLOSS-ID, PAD-ID\}$), where $GLOSS-ID$ denotes the ID number of the sign gloss in the vocabulary and $PAD-ID$ denotes the ID number corresponding to the space. Briefly, $Vg_{ctc}$ is a sequence of $P$ optimized by CTC layer. The lengths of $P$ and $V_{f}$ are equal and one-to-one, while the length of $Vg_{ctc}$ is relatively much smaller. Our final goal is to obtain the alignment between the visual nodes in the video frame sequence $V_{f}$ and the textual nodes in the original sign gloss sequence $P$, denoted as $AlignArr=\{A(V_{g0}),A(V_{g1}),... ,A(V_{gi}),... ,A(V_{gL})\}$, where $V_{gi}$ is a sequence of textual nodes defined differently from $Vg_{ctc}$, which is a sequence further generated by the pseudo-label sequence $P$; $A(V_{gi})$ is a binary array, for example, $A(V_{gi})=\{V_{gi},V_{fj}\}=\{V_{fj},P_{j}\}$ ($i$ is generally not equal to $j$) indicates that there is a graph alignment relationship between $V_{fi}$ and $P_{j}$; $L$ denotes the total number of binary arrays. In general, $L$ is less than the number of $V_{f}$ frames $N$. In the next section, we will describe how to calculate $V_{gi}$ and $AlignArr$.

\begin{figure}[h]
  \centering
  \includegraphics[width=\linewidth]{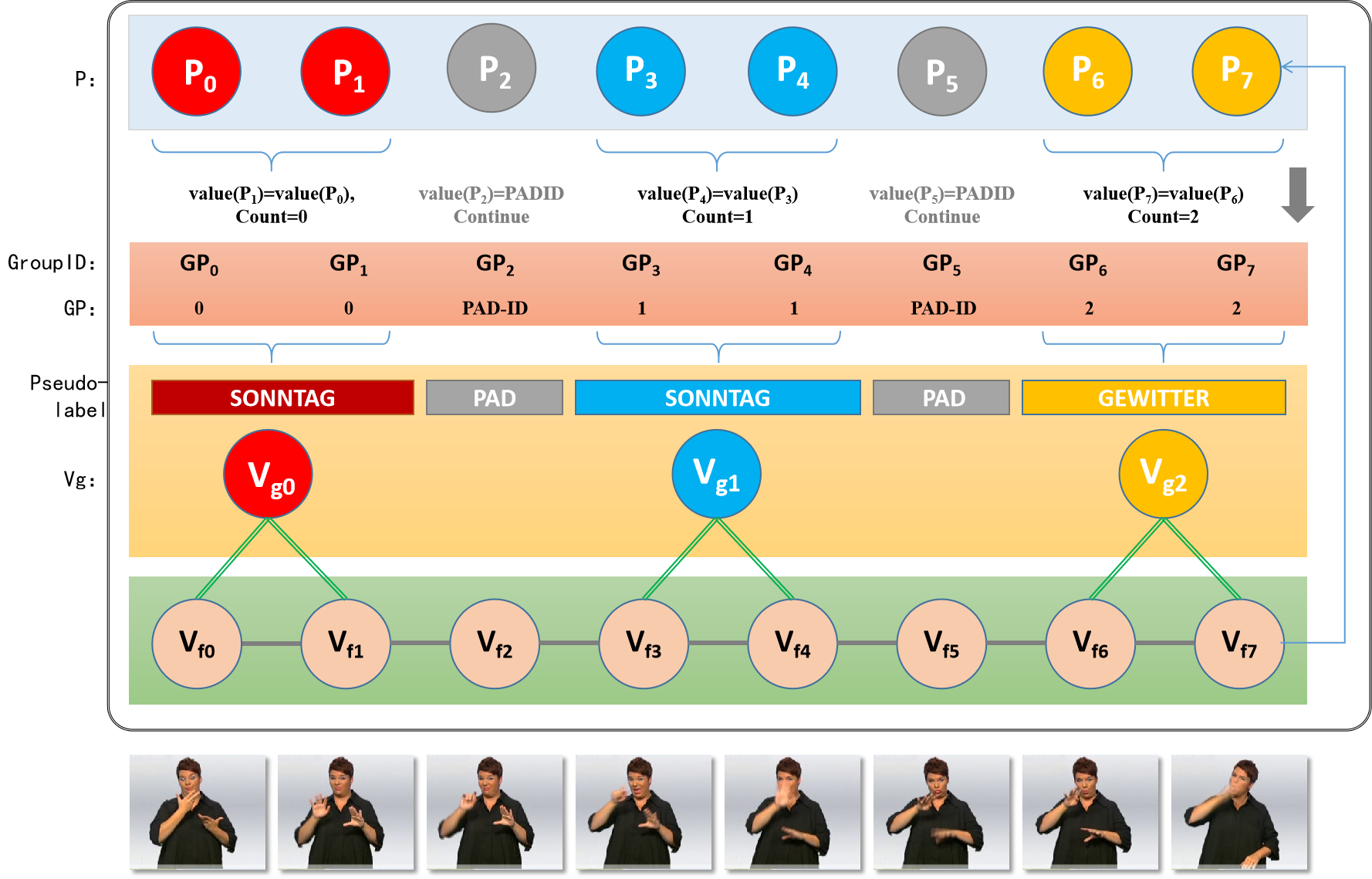}
  \caption{A running example of aligning graph network nodes.}
\end{figure}

\subsubsection{Graph Node Sequence Alignment}
Based on the $P$ obtained above, we can calculate the mapping relationship array $AlignArr$ between the visual nodes in $V_f$ and the textual nodes in $V_g$ as follows:

Initialize a counter $COUNT=-1$, whose value indicates the grouping index value of a valid node, and then iterate through all elements in $P$ in an orderly manner, where $P_i$ indicates the i-th $P$ element. $i==0$ is a special case, and we specify this $P_i$ as a valid node by default; In the case of $i>0$, if $P_{i}==PAD-ID$, which means a space, then skip $P_i$, but this $P_i$ is an invalid node (invalid node has no grouping index value); And if $P_{i} \not= PAD-ID$ and $P_{i} \not= P_{i-1}$, then this $P_i$ node is a valid node, and the $COUNT$ value is automatically increased by 1, whose value indicates the grouping index of $P_i$ and is denoted as $GP_{i}$; And if $P_{i} \not= PAD-ID$ but $P_{i} == P_{i-1}$, it means that this $P_{i}$ node is also a valid node, and the $COUNT$ value remains unchanged.

After traversing the $P$-array, for the sake of description, we merge the neighboring valid nodes with the same grouping $GP$ value to get a new sequence of nodes $V_g$ and number them sequentially. For example, $V_{gj}$ corresponds to the $P$-node with grouped $GP$ value $j$, where the $V_{f}$-node mapped by $V_{gj}$ is the same as the $V_f$-node mapped by the corresponding $P$-node. Finally, we add the $V_g$ values and their mapped $V_f$ values as a binary array to the $AlignArr$ array to get our desired set of visual-textual alignment relations. For example, a binary array of $AlignArr[i]$ is $\{V_{g2},V_{f3}\}$, which indicates that the visual node $V_{f3}$ and the textual node $V_{g2}$ have a graph-connected mapping relationship, as Fig. 3 shows a running example.

With the node sequence alignment algorithm described above, we further mined the deep semantic information of the pre-trained CSLR model, made full use of the clustering relationship of the internal nodes of the pseudo-label sequence, implemented sign language segmentation in a semi-supervised manner, and initially mapped the textual and visual node relationships for graph connectivity and feature fusion.

\subsection{Multi-modal Graph Encoder Construction}
In essence, our multi-modal encoder can be considered as a multi-modal extension of a graph neural network. To construct the encoder, we stack multiple \textbf{Multi-Modal Fusion Layers} based on text-image alignment to learn node feature representations, which provide the attention mechanism-based context vectors for the subsequent NMT module. Next, we describe in detail the multi-modal graph initialization and the construction of a multi-modal graph encoder to obtain multi-modal embedding layers.

\subsubsection{Multi-modal Graph Initialization}
Unlike text-based NMT models or video-only based SLT models, the input to our model requires not only \textbf{Spatial Embeddings} to represent the video frame nodes, but also \textbf{Word Embeddings} to represent the sign glosses in the source sequence and the spoken translations in the target sequence. Given a sign language video frame node $V_{ft}$, the pre-trained convolutional module learns to extract nonlinear frame-level spatial feature representation as:
\begin{equation}
  O_{t} = SpatialEmbedding(V_{ft})=CSLR_{CNN}(V_{ft}).
\end{equation}

And for \textbf{Word Embeddings}\cite{zheng2022using}, we represent the pseudo-sign language gloss sequence nodes $V_{gu}$ as one-hot vectors that linearly map to a more dense space as:
\begin{equation}
  X_{u} = WordEmbedding(V_{gu}).
\end{equation}

Since the visual node representation and the textual node representation are mapped into the same latent space, the multi-modal graph features can be fed to the multi-modal encoder. Given that our multi-modal encoder is designed based on a fully self-attention mechanism and lacks positional representation within the sequence, we will use Positional Encoding \cite{vaswani2017attention} to represent the temporal information, which generates a unique vector in the form of a phase-shifted sine wave for each time step as follows:
\begin{equation}
  O_{t} = O_{t} +PositionalEncoding(t),
\end{equation}

\begin{equation}
  X_{t} = O_{u} +PositionalEncoding(u).
\end{equation}

\begin{figure}[h]
  \centering
  \includegraphics[width=\linewidth]{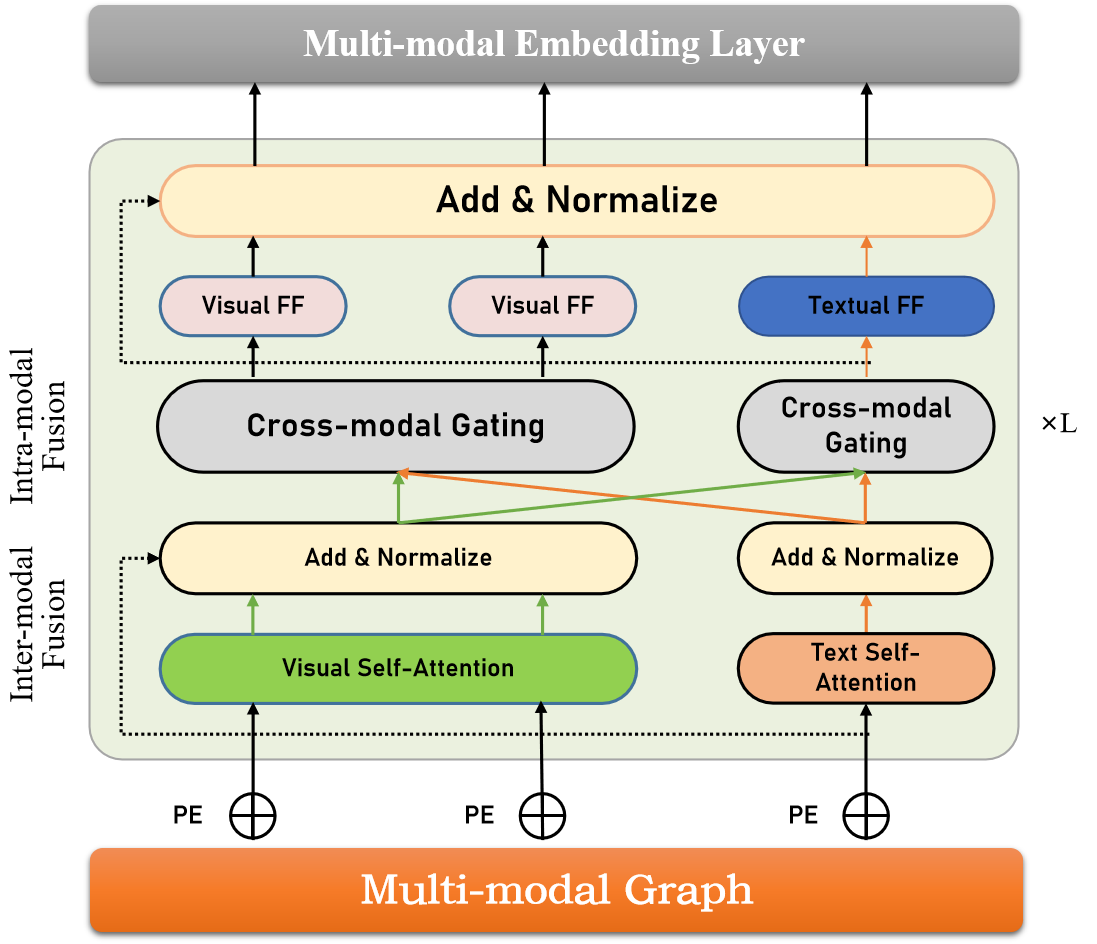}
  \caption{Schematic diagram of multi-modal graph encoder structure.}
\end{figure}

\begin{figure*}[h]
  \centering
  \includegraphics[width=0.8\linewidth]{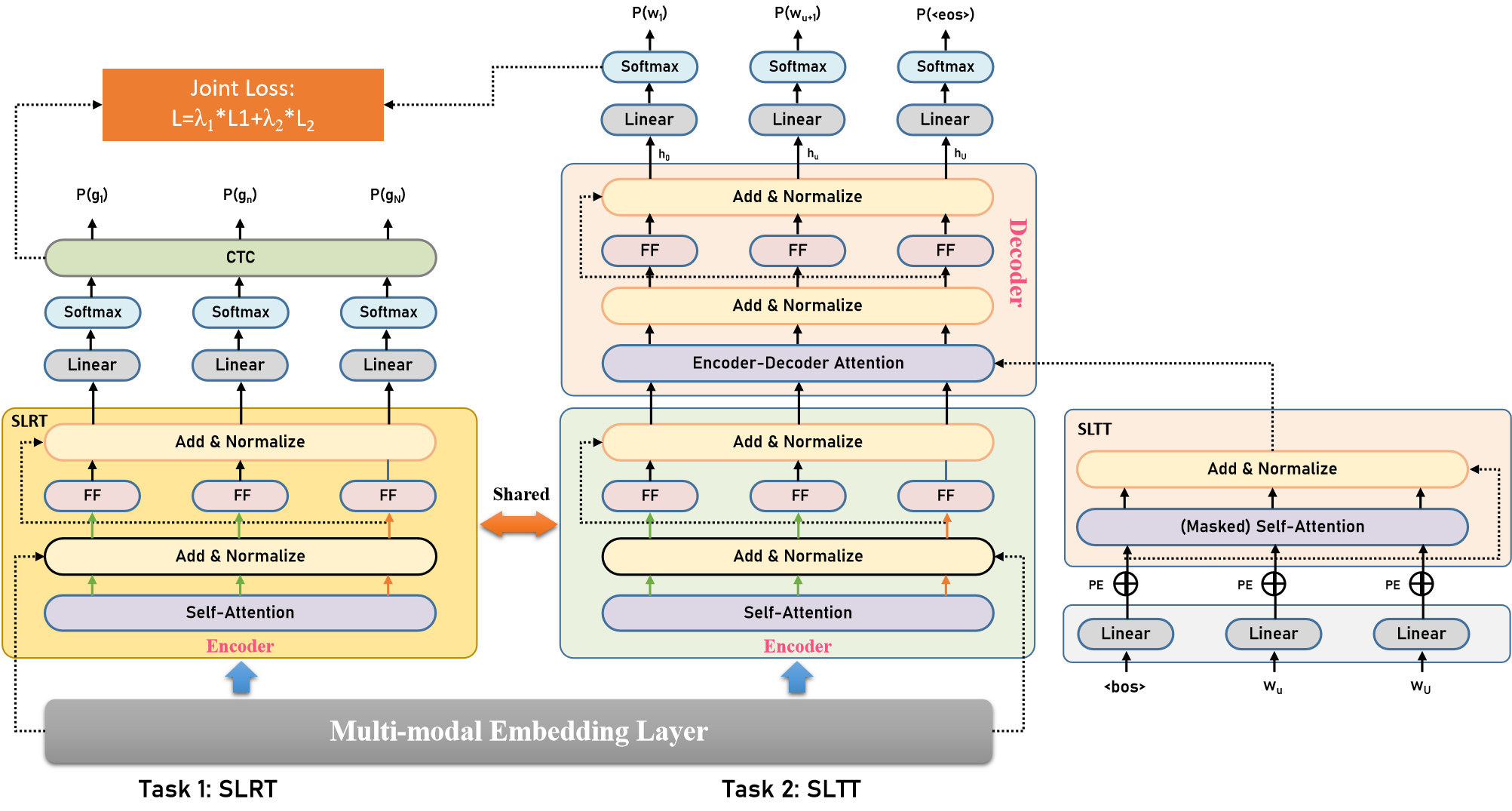}
  \caption{Our multi-task learning machine translation framework based on Transformer.}
\end{figure*}

\subsubsection{Multi-modal Graph Encoder}
Fig. 4 shows the stacked multi-modal fusion layers to encode the above multi-modal graph. In each fusion layer, we perform an \textbf{Inter-modal Fusion} and an \textbf{Intra-modal Fusion} sequentially to update all node states. Formally the initial state of each textual node $v_{xi}$ is denoted as $\{H_{xi}^{0}\}$ and the initial state of each visual node $v_{oj}$ is denoted as $\{H_{oj}^{0}\}$.

In the \textbf{Inter-modal Fusion} step, a contextual representation of each textual node $C_{x}^{(l)}$ or visual node $C_{o}^{(l)}$ is generated by receiving adjacent information within the same modality and using a self-attention mechanism as:
\begin{equation}
\begin{split}
  C^{(l)} = MultiHeaad(H^{(l-1)},H^{(l-1)},H^{(l-1)}),
  \end{split}
\end{equation}
where $MultiHead(Q,K,V)$ is a multi-headed self-attention function. $C^{(l)}$ denotes $C_{x}^{(l)}$ or $C_{o}^{(l)}$, and $H^{(l-1)}$ denotes the corresponding $H_{x}^{(l-1)}$ or $H_{o}^{(l-1)}$.

In the \textbf{Intra-modal Fusion} step, a \textbf{Cross-Modal Gating Mechanism} \cite{teney2018tips,kim2018bilinear} is used to integrate the semantic information from textual modality into visual modality, or the semantic information from the visual modality into textual modality, in order to better represent the degree of contextual fusion in each modality:
\begin{equation}
M_{xi}^{(l)}= \sum_{j \in A(v_{xi})} \alpha_{i,j} \cdot C_{oj}^{(l)},
\end{equation}
\begin{equation}
\alpha_{i,j} = Sigmoid(W_{1}^{(l)}C_{xi}^{(l)}+W_{2}^{(l)}C_{oj}^{(l)}).
\end{equation}

where $A(v_{xi})$ is the set of neighboring visual nodes of $v_{xi}$, and $W_{i}^{(l)}$ are the trainable matrices. $M_{oj}^{(l)}$ is calculated similarly to $M_{xi}^{(l)}$.

Finally, a positional feed-forward network FFN is used to generate textual node states and visual node states separately. In practical applications, it is not optimal to use both $H_{x}^{l}$ and $H_{o}^{l}$ together \cite{yin2020novel}. Since the visual features are the main features in the SLT task and $H_{o}^{l}$ has actually incorporated the textual information, we will use only $H_{o}^{l}$.


\subsection{Multi-task Learning Based Joint Machine Translation}
The multi-modal embedded features from \textbf{Multi-modal Graph Encoder} are then fed to the Transformer-based NMT module for joint Seq2Seq learning, as shown in Fig. 5. One sub-module of our multi-task module is \textbf{Sign Language Recognition Transformer} (SLRT), which learns to recognize multi-modal sign language features into corresponding sign gloss sequences. And another sub-module is \textbf{Sign Language Translation Transformer} (SLTT), which learns to translate multi-modal sign language features into corresponding spoken translation sentences. Since SLRT and SLTT share the same Encoder, SLTT can better utilize SLRT as an intermediate supervision and take the spatio-temporal representation that SLRT has learned as contextual input.

Formally, given a sign language video sequence with $T$ frames, denoted as $I=(I_{1},... ,I_{T})$, SLRT and SLTT then are trained to learn and maximize the conditional probabilities $p(G|I)$ and $p(S|I)$, respectively, where $p(G|I)$ denotes the conditional probability of generating a sign gloss sequence $G=(g_{1},...,g_{N})$ with $N$ words, and $p(S|I)$ denotes the conditional probability of generating a spoken translation sequence $S=(w_{1},... ,w_{U})$ with $U$ words. After inputting the multi-modal graph embedding sequences, the SLRT encoding process is expressed as follows:
\begin{equation}
  z = SLRT(MultiEmd),
\end{equation}

where $z$ denotes the spatio-temporal feature representation of the multi-modal graph embedding features.

Then, we use a linear mapping layer and a softmax activation function to obtain the probability $p(gt|I)$ of generating the sign glosses. After that, the conditional probability $p(G|I)$ is computed by marginalizing all possible alignments of $I$ to $G$ as:
\begin{equation}
p(G|I) = \sum_{\pi \in B} p(\pi|I),
\end{equation}

where $\pi$ is the path and $B$ is the set of all feasible paths corresponding to $G$.

Finally, the CTC Loss \cite{graves2006connectionist} of the SLRT module is calculated as:
\begin{equation}
L_{R} = 1-p(G^{*}|I),
\end{equation}

where $G*$ is the reference sequence of the sign glosses.
Similar to the SLRT, the SLTT decoding process is denoted as:
\begin{equation}
h_{u+1} = SLTT(MultiEmb_{u}|MultiEmb,z)
\end{equation}

The SLTT is trained by decomposing the sequence-level conditional probabilities $p(S|I)$ into ordered conditional probabilities:
\begin{equation}
p(S|I)=\prod_{u=1}^{U} p(w_{u}|h_{u}).
\end{equation}

Finally, we update our network parameters by minimizing the joint loss $L$, where $L$ is the weighted sum of the loss $L_R$ for SLRT and the loss $L_T$ for SLTT as:
\begin{equation}
L = \lambda_{R}L_{R}+\lambda_{T}L_{T},
\end{equation}

where $\lambda_{R}$ and $\lambda_{T}$ denote the hyper-parameters of the weights.

\section{Implementation Details}
\subsection{Dataset}
We evaluate the network performance using the RWTH-PHOENIX-Weather 2014T (PHOENIX14T) dataset \cite{camgoz2018neural}, a large-vocabulary German SLT parallel corpus, which is the first publicly available SLT dataset, and the only that currently available at a larger scale. The PHOENIX14T includes sign language videos, sign gloss annotations, and German text translations. Since PHOENIX14T is extended from PHOENIX14 corpus, the primary benchmark database for CSLR in recent years, this makes it the only dataset applicable for training and evaluating joint training techniques for CSLR and SLT.

\subsection{Evaluation Metrics}
Our multi-tasking architecture involves two tasks, i.e., SLT and CSLR, and the evaluation metrics used for these two tasks are different. Since the main goal of our model is the translation task, we use the evaluation metrics of the SLT task as the primary measure. BLEU \cite{papineni2002bleu} and ROUGE \cite{lin2004rouge} are the most common metrics in the field of NMT, so we use them to evaluate the neural SLT. Note that our BLEU scores use four different criteria, i.e., BLEU-1, BLEU-2, BLEU-3, and BLEU-4 (n-gram from 1 to 4), respectively, while ROUGE score refers to ROUGE-L F1-SCORE. During training stage, we filter the optimal model configuration based on the BLEU-4 scores on the development set.

In addition, to measure the performance of the CSLR module, we use the Word Error Rate (WER), which represents the minimum distance to convert the generated prediction sequence into the corresponding reference sequence.

\subsection{Task Protocols}
We follow the evaluation protocol and task naming conventions for the PHOENIX14T dataset as specified in \cite{camgoz2018neural} and \cite{camgoz2020sign}: \textbf{Sign2Text} is the ultimate goal of SLT, translating continuous sign language videos into spoken translations without any intermediate representation; \textbf{Gloss2Text} is a text-to-text NMT task that aims to translate sign gloss annotations into corresponding spoken translations; \textbf{Sign2Gloss2Text} first uses a pre-trained CSLR model to extract sign gloss sequences, and then trains the \textbf{Gloss2Text} network by using the predicted sequences; \textbf{Sign2Gloss→Gloss2Text} is similar to \textbf{Sign2Gloss2Text}, but uses a pre-trained \textbf{Gloss2Text} network to translate the intermediate sign gloss sequences from the CSLR model into spoken translations; \textbf{Sign2Gloss} essentially performs the CSLR protocol; \textbf{Sign2(Gloss+Text)} denotes the task mode based on a multi-task learning framework, which learns both CSLR and SLT through joint training.

In addition to evaluating the model based on the above protocols, we follow the same naming convention and introduce a new protocol, i.e., \textbf{GSign2(Gloss+Text)}, which denotes the multi-task learning mode performed by our proposed graph neural SLT model.

\begin{table*}
  \caption{Comparison of our models with various benchmark models.}
  \label{tab:commands}
  \scalebox{0.63}{
  \begin{tabular}{ccrcccccccccc}
    \hline
    & & & \multicolumn{5}{c}{DEV SET} & \multicolumn{5}{c}{TEST SET}\\
    
    & \# & MODEL & WER &	BLEU-1 & 	BLEU-2	& BLEU-3	& BLEU-4 & WER &	BLEU-1 & 	BLEU-2	& BLEU-3	& BLEU-4\\
    \hline
    \multicolumn{13}{c}{\textbf{GROUP 1: CSLR}}\\
    \multirow{2}{*}{G1} & 1a & CNN+LSTM+HMM \citeyearpar{koller2019weakly}	& 24.50	& -	& -	& -	& - & 26.50	& -	& -	& -	& -\\
      & 1b & Sign2Gloss \citeyearpar{camgoz2020sign}	& 24.88	& -	& -	& -	& -& 24.59	& -	& -	& -	& -\\
      
     \hline
    \multicolumn{13}{c}{\textbf{GROUP 2: Multistage SLT}}\\
    \multirow{2}{*}{G2} & 2a & Sign2Gloss→Gloss2Text \citeyearpar{camgoz2018neural}	& -	& 41.08	& 29.10	& 22.16	& 17.86 & -	& 41.54	& 29.52	& 22.24	& 17.79\\
      & 2b & Sign2Gloss→Gloss2Text \citeyearpar{camgoz2020sign}	& -	& 47.84	& 34.65	& 26.88	& 21.84 & -	& 47.74	& 34.37	& 26.55	& 21.59\\
    
    \hline
    \multicolumn{13}{c}{\textbf{GROUP 3: End-To-End SLT}}\\
    \multirow{3}{*}{G3-1} & 3a & Sign2Text \citeyearpar{camgoz2018neural}	& -	& 31.87	& 19.11	& 13.16 & 9.94 & -	& 32.24	& 19.03	& 12.83	& 9.58\\
      & 3b & Keypoint Based \cite{ko2019neural}	& -	& 30.34	& 18.06	& 12.62	& 9.67 & -	& 29.18	& 17.17	& 11.87	& 9.01\\
      & 3c & Domain Adaptation \citeyearpar{orbay2020neural}	& -	& -	& -	& -	& - & -	& 34.84	& 22.07	& 15.75	& 12.21\\  
      \hline
    \multirow{2}{*}{G3-2} & 3d & Multi Task(Vision-Based) \citeyearpar{orbay2020neural} & -	& -	& -	& -	& - 	& -	& 37.11	& 24.10	& 17.46	& 13.50\\
	& 3e	& Multi Task(DeepHand) \citeyearpar{orbay2020neural} & -	& -	& -	& -	& - & -	& 38.50	& 25.64	& 18.59	& 14.56\\
	
    G3-3 & 3f	& Sign2Text \citeyearpar{camgoz2018neural}	& -	& 45.54	& 32.60	& 25.30	& 20.69 & -	& 45.34	& 32.31	& 24.83	& 20.17\\
    \hline
    \multirow{2}{*}{G3-4}	& 3g	& Best Recog. Sign2(Gloss+Text) \citeyearpar{camgoz2020sign}	& 24.61	& 46.56	& 34.03	& 26.83	& 22.12 & 24.49	& 47.20	& 34.46	& 26.75	& 21.80\\
	& 3h &	Best Trans. Sign2(Gloss+Text) \citeyearpar{camgoz2020sign} & 24.98	& 47.26	& 34.40	& 27.05	& 22.38 	& 26.16	& 46.61	& 33.73	& 26.19	& 21.32\\
	\hline
    \multirow{2}{*}{G3-5}  & 3i	& Our GSign2(Gloss+Text)	& 24.28	& 47.29	& 34.62	& 27.23	& 22.41 & 24.37	& 47.20	& 34.45	& 26.99	& 22.08\\

	& 3j	& Our GSign2(Gloss+Text) (+Iteration)	& \textbf{24.02}	& \textbf{47.40}	& \textbf{34.75}	& \textbf{27.31}	& \textbf{22.44} & \textbf{24.03}	& \textbf{47.26}	& \textbf{34.57}	& \textbf{27.17}	& \textbf{22.36}\\
     \hline
  \end{tabular}
  }
\end{table*}

\subsection{Experimental Configurations}
\textbf{Basic Setup}: we used the same framework as the baseline model; used the PyTorch framework \cite{paszke2017automatic} for all network components except for the Beam Search for CTC decoding, which was implemented using TensorFlow \cite{abadi2016tensorflow}; used Xavier \cite{glorot2010understanding} to initialize and train networks from scratch; used Dropout with a drop probability of 0.1 on the self-attention and word embedding layers; trained the network using the Adam \cite{kingma2014adam} optimizer. Batch size was set to 32, Learning Rate was set to $10^{-3} (\beta_{1}=0.9,\beta_{2}=0.998)$, and Weight Decay was set to $10^{-3}$;Transformer consists of 512 hidden units and 8 heads per layer. The performance of the development set was tracked to roughly reflect the model training level. During the training and validation, Greedy Search was used to decode. And during the inference, Beam Search was used to decode. We also used a length penalty strategy \cite{wu2016google} with $\alpha$ values ranging from 0 to 2.

\textbf{Spatial Embedding Layer Setup}: any CNN module can be used as a spatial embedding layer, but due to hardware constraints (e.g., GPUs), we will use pre-trained CNNs as spatial embeddings. Based on configuration experiment of the spatial embedding \cite{camgoz2020sign}, we used Batch Normalization \cite{ioffe2015batch} and ReLU \cite{nair2010rectified} to normalize the input sequence features and allow the network to learn more abstract nonlinear representations.

\textbf{Transformer Layer Setup}: Different Transformer layers have different effects on the model. Having more layers allows the network to learn more abstract representations, but it can also easily lead to over-fitting, especially when there is less data. Referring to the parameter experiment of \cite{camgoz2020sign}, we found that the Transformer performs best when the number of layers is set to 3

\textbf{Loss Weight Assignment Setup}: The joint loss $L$ is the weighted sum of $L_{R}$ and $L_{T}$. When the weight $\lambda_{R}$ or $\lambda_{T}$ is 0, the \textbf{Sign2Text} or \textbf{Sign2Gloss} protocols are performed, respectively. But when both weights are set to non-zero values, the recognition and translation tasks are performed jointly. Based on the results of configuration experiment \cite{camgoz2020sign}, we used $\lambda_{R}=5.0$ and $\lambda_{T}=1.0$.

\section{Experiments and Analysis}
\subsection{Baseline}
We used the Transformer-based SLT model proposed by Camgoz et al \cite{camgoz2020sign} as the baseline model, which is the state-of-the-art model. The model uses the convolutional network module from the pre-trained CSLR model as the spatial embedding to extract the spatial features of sign language, and then performs the CSLR task and the SLT task by the Transformer encoder and CTC layer combination and the Transformer encoder-decoder combination, respectively, in an end-to-end joint training manner. Since both task modules share the same encoder, the CSLR module can improve the ability of encoder to learn spatio-temporal features by back-propagation optimization.

\subsection{Quantitative Results}

As shown in Table 1, we compared our proposed graph SLT model (\textit{G3-5}) with a variety of the most popular models currently available on a variety of tasks. Compared with the state-of-the-art SLT baselines (\textit{G3-4}), our optimal model improves by about +0.6 BLEU-4 score relatively.

For the convenience of multi-aspect comparison, the models in Table 1 are divided into three groups::

\textbf{Group 1} includes the currently popular CSLR systems. Since our model is based on a multi-task learning mode, the performance of the auxiliary CSLR module is also an important metric to measure the performance laterally. In terms of the performance of the CSLR module, compared with the CSLR systems \textbf{1a} and \textbf{1b} in \textit{G1}, our model reduced 2.40 and 0.52 WER scores, respectively (the smaller the WER score, the better the performance), where \textbf{1a} is a pure CSLR task while \textbf{1b} is a recognition sub-task based on the baseline model. Although the performance of the recognition task is not our primary metric, the recognition module shares encoder parameters with the translation module, so the comparison with this group also reflects that the encoder of our model learns better representations of spatio-temporal features.

\textbf{Group 2} includes the multi-stage SLT systems. This step-by-step implementation of the SLT systems actually performs CSLR and SLT sequentially, i.e., the pseudo-gloss sequences generated by the CSLR system are used as intermediate transitions and then fed to the NMT system to generate the spoken translations. Compared to this method, our model relatively improved about +0.8 BLEU-4 score. Even compared to other end-to-end models (e.g., \textbf{3g}, \textbf{3h}), we found that the multi-stage SLT model is no longer advantageous. In terms of both effectiveness and efficiency, our model with an end-to-end manner improves considerably over the multi-stage manner.

\begin{table*}
  \caption{Experiments on verifying different strategies in multi-modal graph embedding layer.}
  \label{tab:commands}
  \scalebox{0.72}{
  \begin{tabular}{c|r|cccc|cccc}
    \hline
    & & \multicolumn{4}{c|}{DEV SET} & \multicolumn{4}{c}{TEST SET}\\
    \# & MODEL & 	BLEU-1 & 	BLEU-2	& BLEU-3	& BLEU-4 & 	BLEU-1 & 	BLEU-2	& BLEU-3	& BLEU-4\\
    \hline

    \multirow{2}{*}{1} & Best Recog. Sign2(Gloss+Text) \citeyearpar{camgoz2020sign} 	& 46.56	& 34.03	& 26.83	& 22.12 & 47.20	& 34.46	& 26.75	& 21.80\\
     & Best Trans. Sign2(Gloss+Text) \citeyearpar{camgoz2020sign} & 47.26	& 34.40	& 27.05	& 22.38	& 46.61	& 33.73	& 26.19	& 21.32\\
     \hline
    2 & Our GSign2(Gloss+Text) ($H_{o}^{(l)}$) 	& \textbf{47.29}	& \textbf{34.62}	& \textbf{27.23}  & \textbf{22.41} &	\textbf{47.20}	& \textbf{34.45}	& \textbf{26.99}	& \textbf{22.08}\\
    3 & Our GSign2(Gloss+Text) ($H_{x}^{(l)}$) & 46.00 & 32.93 & 25.46 & 20.67 & 45.91	& 32.89	& 25.38	& 20.58\\
    4 & Our GSign2(Gloss+Text) ($H_{o}^{(l)}+H_{x}^{(l)}$) &46.63 &33.68 &26.17 &	21.32 & 46.34	& 33.42	& 25.92	& 21.09\\
    \hline
  \end{tabular}
  }
\end{table*}

\textbf{Group 3} includes the end-to-end SLT systems. The core structures in this group all use a sequential structure of spatial embedding modules (CNN networks) and NMT modules. For further comparison, we subdivide the experiments into multiple groups: Group \textbf{G3-1} indicates that the CNN network of the model is not processed with a pre-trained sequential CSLR model, and its CNN part is only initialized with models from other non-sign language datasets; Unlike the group \textbf{G3-1}, the models in group \textbf{G3-2} introduce a multi-task learning framework to optimize CNNs. Note that, in contrast to the multi-task learning mode in group \textbf{G3-4}, the learning goal of the multi-task in \textbf{G3-2} is to enhance the CNN part only; Group \textbf{G3-3}, on the other hand, uses a pre-trained CSLR system to pre-process the CNN part. The comparison of these 3 groups in \textbf{Group 3} further demonstrates the importance of the spatial embedding layer to the model, indicating that the current best practice is to use the pre-trained CSLR model for extracting the spatial embedding features.

The \textbf{G3-4} group applies a different training mode by using a multi-task learning method combining recognition and translation to improve the NMT module. This is the most advanced architecture publicly available, which enhances the spatio-temporal representation in a superior performance. Compared with gropu \textbf{G3-4}, our models of \textbf{G3-5} are further improved in the representation of spatio-temporal features. In addition, by comparing model \textbf{3i} with the models in group \textbf{G3-4}, it can been see that our multi-modal graph encoder structure has better optimization effect. This clearly illustrates the importance of using a multi-modal graph fused embedding layer, which has a better feature representation capability than the spatial embedding layer of a single mode. From this we can infer that our proposed graph-based encoder effectively incorporates the external textual information of sign language: 1) The sign language segmentation implemented using a weakly supervised approach effectively aligns textual and visual information using graph structure and enables the fusion of multi-modal features; 2) The introduction of sign language lexical-level annotations into the encoder as external textual knowledge can help the model obtain semantic information that cannot be directly mined by common deep networks.

Furthermore, compared within the group \textbf{G3-5} (\textbf{3i} vs. \textbf{3j}), model \textbf{3j} improved by about +0.3 BLEU-4 score, benefiting from the dynamic iterative alignment method \textbf{3j} using pseudo-labels during training (2 iterations). However, in terms of efficiency, \textbf{3j} is slightly slower and consume more memory than \textbf{3i}, due to the process of reconstructing the graph network and judging iterations.

\subsection{Ablation Analysis}
In the process of constructing the multi-modal graph embedding layer in the graph network module, we used a Cross-Modal Gating Mechanism to generate the visual node states $H_{o}^{(l)}$ and the textual node states $H_{x}^{(l)}$, respectively. But in the practice, only $H_{o}^{(l)}$ was utilized. We believe that $H_{o}^{(l)}$ has incorporated the textual information in the visual information, while using only the textual node states $H_{x}^{(l)}$ or using the fused features of both $H_{o}^{(l)}$ and $H_{x}^{(l)}$ has worse performance. To validate the hypothesis, we conducted this ablation experiment.

In this experiment, in order to simplify the operation and reduce the complexity, our graph neural SLT model did not use the step of iterative graph construction. From Table 2, we can see that the experimental results vary widely when using different cross-modal fusion features as the multi-modal graph embedding layer. For our proposed method, the sign language video features remain as the primary input features, while the sign glossed are used as a secondary textual feature. In other words, the effectiveness of our model depends mainly on the extraction ability of visual features. Although $H_{x}^{(l)}$ also integrates visual features into the textual information, it weakens the expression of visual features, so it is not difficult to infer that \#3 performs worse than \#2. As a comparison, \#4 fused $H_{x}^{(l)}$ and $H_{o}^{(l)}$, which still performed worse than \#2, but better than \#2. We speculate that the main reason for the poorer fusion is that $H_{x}^{(l)}$ interferes with $H_{o}^{(l)}$ feature expression and introduces more noise. This experiment also indirectly shows that visual features, as the main features of multi-mode, still convey the most core information. Although textual features have facilitating role, they cannot be used as the primary feature representation, otherwise the model performance will be weakened.

\section{Conclusions and Future Work}
In this work, we proposed a dynamic graph-based neural SLT model with multi-modal feature fusion using semantic knowledge. We applied the semantic segmentation of sign language, graph networks, and multi-modal feature representation to the SLT model for the first time, overcoming various challenges of modeling. The model performance was improved by introducing knowledge of external sign language linguistics to deeply explore the implicit language features possessed by sign languages. To validate the effectiveness of our model, we conducted multiple sets of experiments in the benchmark database PHOENIX-2014T. The experimental results demonstrate the effectiveness and advancement of our proposed neural networks.

In future work, we propose to explore a lightweight graph-based neural SLT model with lower complexity.

\section*{Acknowledgements}
We thank all the reviewers for their helpful comments and suggestions.

\bibliography{anthology,custom}
\bibliographystyle{acl_natbib}

\appendix



\end{document}